\documentclass[10pt,twocolumn,letterpaper]{article}

\usepackage{cvpr}    

\usepackage{multirow}
\usepackage{hyperref}

\usepackage{tikz}
\usetikzlibrary{shapes.geometric, arrows, positioning, fit}
\usepackage{graphicx}
\usepackage{adjustbox}
\usepackage{flushend}

\begin{document}

\title{DMAT: An End-to-End Framework for Joint Atmospheric Turbulence Mitigation and Object Detection}
\author{Paul Hill, Zhiming Liu, Alin Achim, Dave Bull, and Nantheera Anantrasirichai \\
  {Visual Information Laboratory, University of Bristol}, {Bristol}, {UK}}

\maketitle

\begin{abstract}
Atmospheric Turbulence (AT) degrades the clarity and accuracy of surveillance imagery, posing challenges  not only for visualization quality but also for object classification and scene tracking. Deep learning-based methods have been proposed to improve visual quality, but spatio-temporal distortions remain a significant issue. Although deep learning-based object detection performs well under normal conditions, it struggles to operate effectively on sequences distorted by atmospheric turbulence.
In this paper, we propose a novel framework that learns to compensate for distorted features while simultaneously improving visualization and object detection. This end-to-end training strategy leverages and exchanges knowledge of low-level distorted features in the AT mitigator with semantic features extracted in the object detector. Specifically, in the AT mitigator a 3D Mamba-based structure is used to handle the spatio-temporal displacements and blurring caused by turbulence.  Optimization is achieved through back-propagation in both the AT mitigator and object detector.  Our proposed DMAT outperforms state-of-the-art AT mitigation and object detection systems up to a 15\% improvement on datasets corrupted by generated turbulence. The code is available at \url{https://github.com/pui-nantheera/DMAT} and datasets are available at \url{https://zenodo.org/records/17673509}.

\end{abstract}

\section{Introduction}
\label{sec:intro}

When the temperature difference between the ground and the air increases, the thickness of each air layer diminishes, and they ascend rapidly. This leads to rapid  micro-scale changes in the air’s refractive index that degrade the visual quality of video signals and in turn impacting  the effectiveness of automated recognition and tracking algorithms. Objects behind the distorting layers become nearly impossible to recognize using methods designed to operate in non-distorted environments.
\begin{figure}[t]
    \centering
    \includegraphics[trim={3.5mm 4mm 5mm 0},clip, width=\linewidth]{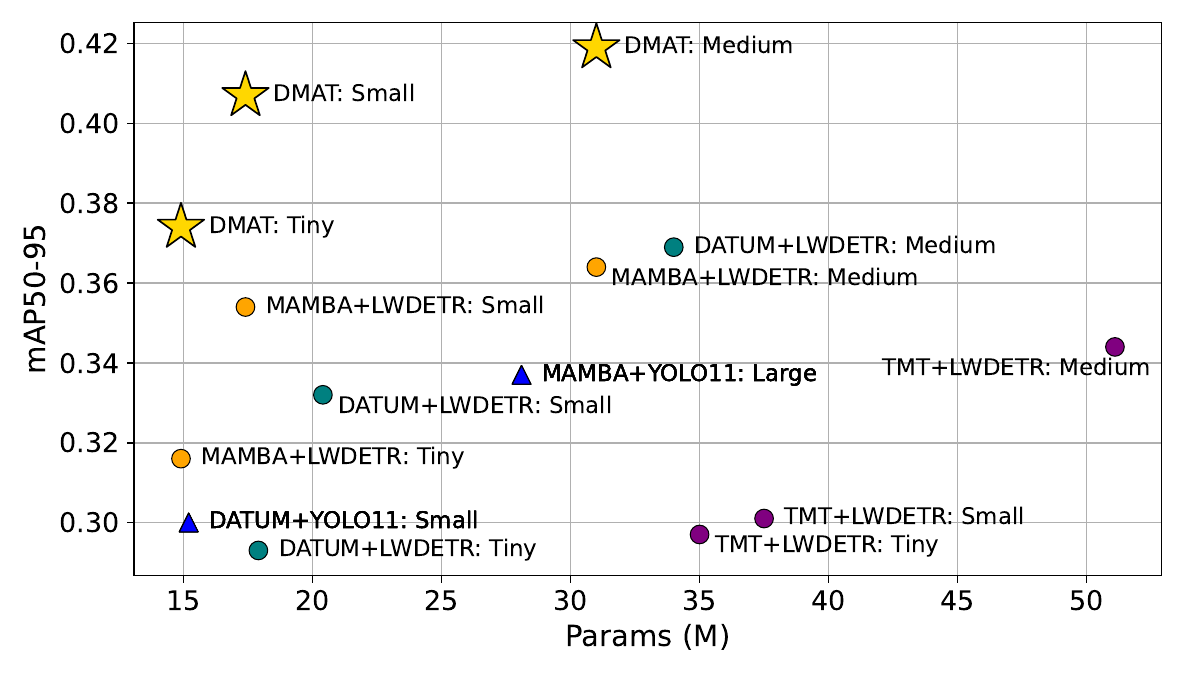}
    \caption{Object detection results  for all object sizes and classes of our proposed method (DMAT) compared to individual AT-mitigation and object detection methods.}
    \label{fig:adresults}
\end{figure}
To address the challenges posed by atmospheric turbulence in video processing, current learning-based approaches typically adopt a variety of strategies. Some methods focus on enhancing visual quality before applying object detection to the restored outputs \cite{Mao2022single,hill2025mamat}. Others retrain existing object detectors on synthetically generated turbulent datasets \cite{Uzun2025augmenting}. JDATT \cite{liu2025jdatt} employs an alternative strategy by jointly training the restoration model and object detector, but this can result in suboptimal performance due to competing objectives. Another strategy involves incorporating motion maps—representing displacements caused by spatial distortions—into the input of object detectors \cite{qin2025unsupervised}. Alternatively, some approaches modify the detector architectures themselves to better handle turbulence-induced distortions \cite{Lau:ATFaceGAN:2020, Hu:Object:2023}. In contrast, we propose a novel approach that aggregates information to enhance feature quality, enabling simultaneous video restoration and object detection (including both classification and localization) within a unified, hybrid framework.

Existing methods have achieved considerable success in mitigating atmospheric turbulence; however, they often fail to ensure that restored or enhanced imagery is optimally configured for automatic object detection. A key challenge is that techniques effective in reducing wavy or ripple effects may inadvertently remove crucial features necessary for object identification within video sequences. To address this, this paper proposes a solution called DMAT (\textbf{D}etection and \textbf{M}itigation of \textbf{A}tmospheric \textbf{T}urbulence), that not only enhances visual quality but also improves object detection accuracy, as demonstrated in Fig.~\ref{fig:adresults}, where our combined architecture outperforms individual AT-mitigation and object detection methods. Our approach aims to strike a balance between video enhancement and applicability in surveillance systems, achieving strong results in both domains.

In this paper, we introduce a novel method that leverages 3D deformable convolutions and Structured State Space sequence (S4) models, known as Mamba. We chose Mamba for its efficiency and ability to model long-range dependencies, and compared to 3D transformers, it yields better PSNR and mAP with fewer parameters as reported in \cite{hill2025deep}.  Our DMAT addresses spatial displacement across the temporal domain, mitigates distortions such as blur, and enhances contrast. It is specifically designed to restore images affected by atmospheric turbulence and subsequently feed them into an object detector, which outputs object locations and classifications. Our joint framework is trained end-to-end with carefully designed loss functions, enabling the high-level features from the object detector and the low-level features from the turbulence mitigator to mutually enhance each other, resulting in significant improvements in both visual quality and object detection accuracy.

The main contributions of this paper are summarized as follows.
\begin{itemize}
    \item A first end-to-end architecture for joint AT mitigation and object detection.  We refer to  this combined architecture as DMAT (Detection and Mitigation of Atmospheric Turbulence).
    \item A new optimized  sets of synthetic turbulent videos based on subsets of the COCO and GOT-10k datasets.
    \item Benchmarking the end-to-end system (DMAT) against a range of individually applied mitigation and detection systems.
\end{itemize}

\section{Related work}
\label{sec:relatedwork}

\subsection{Atmospheric turbulence removal} 
Atmospheric turbulence removal is challenging due to its spatio-temporal variability. Traditional methods involve frame selection, registration, fusion, phase alignment, and deblurring~\cite{Anantrasirichai:Atmospheric:2013, Zhu:Removing:2013, Xie:Removing:2016}. Video-based approaches, like optical flow–guided filtering~\cite{Anantrasirichai:Atmospheric:2018}, struggle with artefacts at motion boundaries. ML-based single-image restoration has been explored~\cite{Gao:Atmospheric:2019}, while early deep learning methods include CNNs for deblurring~\cite{Nieuwenhuizen:deep:2019} and GANs for multi-frame restoration~\cite{Chak:Subsampled:2018}, though often limited to static scenes. Physics-inspired models~\cite{jaiswal2023physics, Jiang_2023_CVPR}, complex-valued CNNs~\cite{anantrasirichai2023atmospheric}, and implicit neural representations~\cite{Jiang_2023_CVPR} offer alternatives. Diffusion models show strong results for single images~\cite{nair2023ddpm}, while transformer-based approaches currently lead in video restoration~\cite{Zhang:Image:2024, zou2024deturb, Liu2026RMFAT} as reported by a recent survey in~\cite{hill2025deep}.

\begin{figure}[t]
    \centering
    \includegraphics[width=\columnwidth]{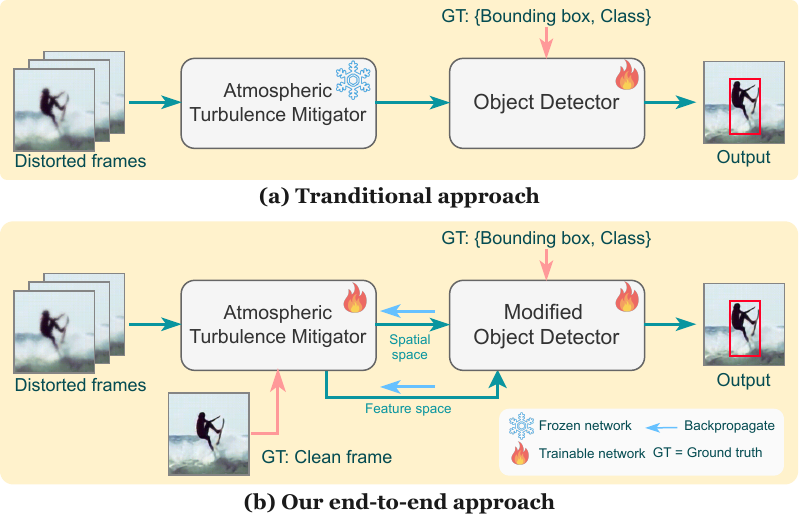}
\caption{Diagrams of (a) transitional approach and (b) our proposed approach for joint restoration and object detection under atmospheric turbulence. }
\label{fig:diagram_brief}
\end{figure}

\begin{figure*}[ht]
\centering
\includegraphics[width=\textwidth]{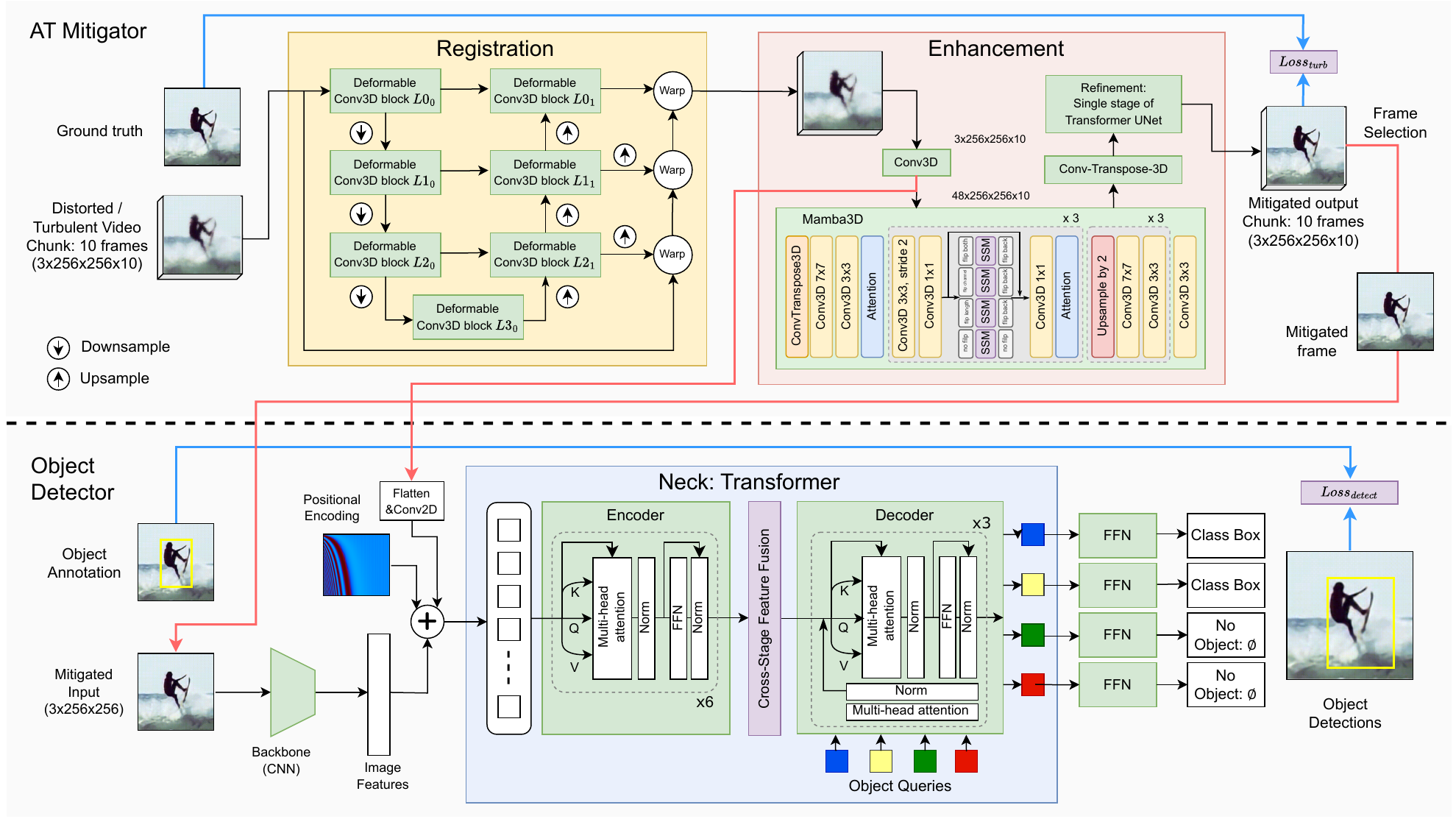}
\caption{Architecture of the proposed DMAT framework for atmospheric distortion mitigation and object detection.}
\label{fig:system_architecture}
\end{figure*}

\subsection{Object Detection} 
Object Detection involves classifying and localizing objects within images. Two-stage detectors, such as R-CNN~\cite{girshick2014rich}, Fast R-CNN~\cite{girshick2015fast}, and Faster R-CNN~\cite{ren2015faster}, first generate region proposals and then classify them, offering high accuracy at the cost of speed. In contrast, one-stage detectors like YOLO~\cite{redmon2016you} perform detection in a single pass, enabling real-time performance. The latest version, YOLOv11~\cite{khanam2024yolov11}, further improves accuracy and speed, though challenges remain with small and occluded objects.
Recently, transformer-based approaches such as DETR~\cite{carion2020end} have emerged, replacing traditional components like anchor boxes and non-maximum suppression with an end-to-end encoder-decoder architecture. Deformable DETR~\cite{zhu2021deformable} introduces multi-scale attention to improve efficiency and small object detection, while DINO~\cite{zhang2022dino} enhances overall accuracy through improved training strategies.

\section{Methodology}
\label{sec:method}

Traditionally, models for AT mitigation and object detection are optimized separately. The AT mitigator is typically trained on synthetic datasets due to the absence of ground truth information. Post-training, the AT mitigator is deployed to generate restored frames that are then utilized to fine-tune the object detector, as illustrated in Fig.~\ref{fig:diagram_brief} (a). In contrast, we propose a joint optimization strategy, as shown in Fig.~\ref{fig:diagram_brief} (b), where high-level features from the object detector  guide the AT mitigator in discerning low-level features with semantic significance. Simultaneously, the object detector benefits from receiving more precise features from the AT mitigator, leading to improved overall performance. This synergistic approach promises to significantly advance the efficacy of both video restoration and object detection in atmospherically challenging conditions.

\subsection{Overview}
The architecture of our proposed framework is depicted in Fig. \ref{fig:system_architecture}. This takes in  a sequence of AT-distorted video frames  and outputs  restored video frames, along with the locations and categories of objects identified within these frames. 
Our framework consists of two primary modules: (1) \textit{AT Mitigator}: This module processes the distorted frames to mitigate the atmospheric effects using a 3D Mamba structure; (2) \textit{Modified Object Detector}: Leveraging a transformer-based model, this module is designed for lightweight and robust object detection. Note that the implementation can be adapted to other detection models as needed. 
The integration of these two modules occurs through (1) direct concatenation and (2) by feeding features extracted after the AT mitigator into the detection head, reducing redundancy in the feature extraction process of the traditional object detector.

\subsection{AT Mitigation}

Our AT mitigation module offers a robust solution to  the problem  of AT distortion, particularly the wavy effect, where there is a time-varying shift of features. Similar to other video processing frameworks, we input a group of frames and treat them as 3D data blocks. Since consecutive frames are highly correlated temporally but spatially shifted due to turbulence, the first step of our method (Registration module) is to register neighboring frames to the current frame. This process is necessary and has been employed with slight variations in both traditional model-based methods~\cite{Anantrasirichai:Atmospheric:2013, Zhu:Removing:2013} and deep-learning-based methods~\cite{Zhang:Image:2024, zou2024deturb}.

\begin{figure} [t!]
    \centering
    \includegraphics[width=0.98\columnwidth]{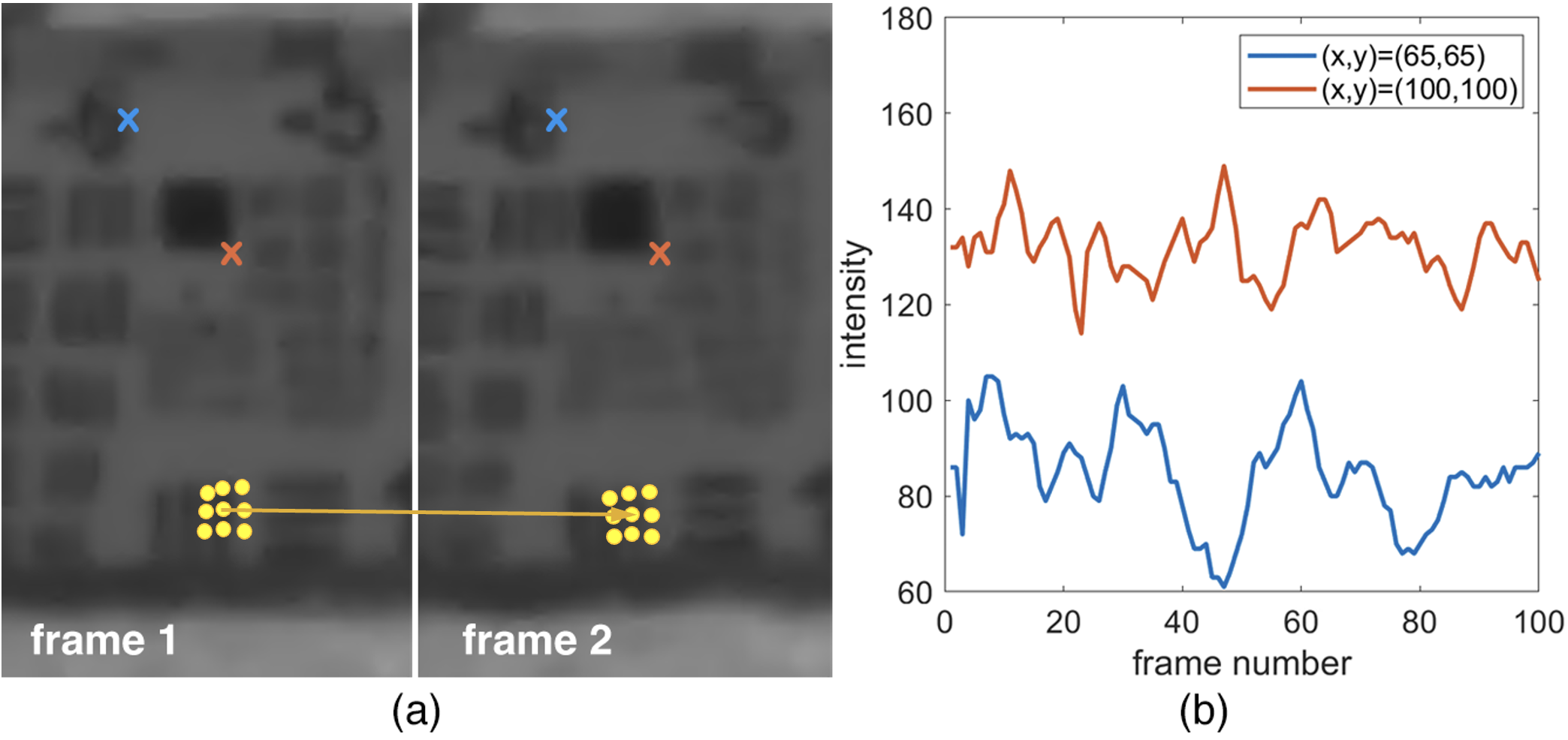}
    \caption{AT distortion. (a) Frames 1 and 2. (b) Temporal intensity variation of two pixels. Yellow dots show the benefit of deformable convolutions.}
    \label{fig:deform}
\end{figure}

\paragraph{Registration module. } We incorporate a UNet-like architecture with deformable 3D convolutions implemented at all scales in both the encoder and decoder, denoted `Deformable Conv3D block' in Fig.~\ref{fig:system_architecture}. This configuration is specifically designed to effectively estimate pixel shifts at different scales across frames. The deformable 3D convolutions enable pixel mappings associated with the current convolution kernel to extend beyond the conventional grid search area, accommodating spatio-temporal variations. This adaptability ensures that identical objects in different frames, even when subjected to varying turbulent distortions, can be aligned to exhibit consistent features as illustrated in Fig. \ref{fig:deform}.

The registration module has a depth scale of 4 ($L0$–$L3$) with kernel sizes of 3$\times$7$\times$7, 3$\times$7$\times$7, 3$\times$5$\times$5, and 3$\times$3$\times$3. Larger kernels are used in the initial layers to provide a wider field of view. The feature spaces expand to 32, 128, 128, and 256, with 3D batch normalization and max pooling for encoder downsamplers, while trilinear interpolation is used for decoder upsamplers. The output of each scale is a motion field which is then used for frame registration. The registered frames are then fed to the enhancement module.

\paragraph{Enhancement module. }
The first step consists in feature extraction using 3D convolutions with a kernel size of 3$\times$7$\times$7. The features are then processed in the 3D Mamba-based UNet-like network and simultaneously fed to the object detector, as shown by the red line in Fig.\ref{fig:system_architecture}. 
We also employ a UNet-like architecture but incorporate 3D Mamba at the encoder, which our studies have shown to outperform the 3D Swin Transformer~\cite{liu2021swin}, in agreement with the findings reported in \cite{hill2025deep}. The Mamba framework is based on the Structured State Space sequence (S4) model. It dynamically adjusts State Space Model (SSM) parameters in response to varying inputs, effectively addressing the common memory and gradient challenges associated with traditional SSM implementations.

Our 3D Mamba-based UNet-like network is based on nnMamba, proposed in~\cite{gong2024nnmamba}, in which the encoder is structured with three layers of residual Mamba blocks (Res-Mamba), while the decoder utilizes a series of three double convolution blocks. To improve efficiency, we integrated an initial 3D convolution before the Mamba module that expands the temporal dimension in order for the 3D Mamba-based UNet-like structure to effectively treat the spatio-temporal tensor as a 3D tensor.

The Res-Mamba block  integrates double convolution layers, skip connections, and a Mamba-in-convolution block that incorporates SSM functionalities between convolutions, as detailed in the following equation:
\begin{equation} F_{\text{o}} = \text{Conv}_{1}(\text{SSM}(\text{Conv}_{1}(F_{\text{i}}))) + \text{Conv}_{1}(F_{\text{i}}), \label{eq:ssminconv} \end{equation}
where $F_{\text{i}}$ and $F_{\text{o}}$ represent the input and output feature maps of the Res-Mamba block, respectively, within a 3D convolution context. $\text{Conv}_{1}(\cdot)$ is executed with a kernel size of 1$\times$1$\times$1, followed by batch normalization and ReLU activation. SSM($\cdot$) is a selective SSM layer, trained using four augmented feature inputs, each flipped and rotated differently~\cite{gong2024nnmamba}. This sophisticated configuration culminates in the application of two Transformer blocks for optimal channel adjustment and refinement of the final output before passing it to the object detector.

\subsection{Object Detection}

The object detector in our framework is based on a Transformer architecture. Features from the mitigated frame, which is the output of the enhancement module, are extracted using a ResNet50 backbone. These features, along with those from the output of the registration process, are flattened and supplemented with positional encoding before being fed into a transformer encoder. The integration of the registration output features ensures that semantic insights from the object detector inform both modules in the AT mitigator, enhancing learning across the entire framework.

Similar to many other transformer-based methods for computer vision tasks, our encoder concatenates six transformer blocks. Each block comprises a multi-head self-attention module, batch normalization, and a feed-forward network (FFN). Inspired by LW-DETR~\cite{chen2024lw} and YOLOv7~\cite{wang2022yolov7}, we employ cross-stage feature fusion to improve gradient flow between the encoder and decoder. Our decoder consists of three transformer decoder blocks, each featuring a self-attention module, a cross-attention module, batch normalization, and an FFN. Finally, the FFNs are used to output the object category, box size and box location. 

We provide three object detection model sizes.  The increased size is reflected in an increase in the number of attention heads, number of transformer layers, embedding dimensions and backbone depth. 

\subsection{Loss function}
\label{sec:lf}

The loss function is computed from two sources: the AT mitigator and the object detector, defined as $Loss_{turb}$ and $Loss_{detect}$, respectively. 
For $Loss_{turb}$, we employ the Charbonnier loss function, which merges the benefits of $\ell_1$ and $\ell_2$ losses, effectively managing outliers in pixel-wise error due to spatial variation from atmospheric turbulence. It is defined as: 
\begin{equation} 
L_\text{Char}(x, y) = \sqrt{(x-y)^2 + \epsilon^2}, \label{Loss} \end{equation} 
where $x$ and $y$ are the predicted and true values, and $\epsilon$ is a small constant (e.g., $1e-3$) to ensure numerical stability. The Charbonnier loss provides a smooth gradient even for small errors, making it ideally suited for addressing subtle discrepancies in turbulence-impacted images.

For $Loss_{detect}$, three loss functions are combined: $L_\text{boxes}$, $L_\text{GIoU}$, and $L_\text{labels}$. $L_\text{boxes}$ is an $\ell_1$ regression loss for box size and location, while $L_\text{GIoU}$ represents the generalized IoU loss \cite{rezatofighi2019generalized}.
For $L_\text{labels}$, we use the Binary Cross-Entropy (BCE) loss to compute the probability $p_\text{ce}$ of class accuracy, using the Intersection over Union (IoU) score. $L_\text{labels}$ is defined as shown in Eq.~\ref{eq:Loss_label} \cite{Cai_2024_BMVC}, 
\begin{equation} 
\begin{split}
L_\text{labels} &= - \frac{1}{N_{pos}+N_{neg}}  [\sum^{N_{pos}} t\log p_\text{ce} \\ 
&+ (1-t) \log (1 - p_\text{ce}) \\
& + \sum^{N_{neg}}  p_\text{ce}^\gamma \log (1 - p_\text{ce})], \: \: t=p_\text{ce}^\alpha \text{IoU}^{(1-\alpha)}
\label{eq:Loss_label} \end{split}\end{equation} 
where $N_{pos}$ and $N_{neg}$ are the numbers of positive and negative boxes, $\gamma$ controls the weight and is set to 2, and $t$ is a smooth transition parameter between a positive target and a negative target, which is set to an exponential decay with $\alpha = 0.25$.



\section{Datasets and implementation}
\label{sec:datasets}

\subsection{Synthetic dataset}

As ground truth data for atmospheric turbulence is unavailable, we generate synthetic AT distortions using the Phase-to-Space (P2S) Transform~\cite{mao2021accelerating}, applying them to both static and dynamic datasets designed for object detection. Due to memory limitations, all images and videos are cropped to a resolution of $256\times256$ pixels.

\begin{table*}[ht]
\centering
\resizebox{0.95\textwidth}{!}{
\begin{tabular}{ll|c|cc|cc|cc|cc}
\toprule
\multirow{3}{*}{\textbf{Mitigator}} & \multirow{3}{*}{\textbf{Detector}} & \multirow{3}{*}{\textbf{Params (M)}} & \multicolumn{6}{c|}{\textbf{Static scenes}} & \multicolumn{2}{c}{\textbf{Dynamic scenes}} \\ \cline{4-11}
& & & \multicolumn{2}{c|}{\textbf{All classes}}  & \multicolumn{2}{c|}{\textbf{Top 10 classes}} & \multicolumn{2}{c|}{\textbf{Car and person}} & \multicolumn{2}{c}{\textbf{Car and person}} \\ \cline{4-11}
& & & \textbf{All sizes} & \textbf{Small} &
\textbf{All sizes} & \textbf{Small} &
\textbf{All sizes} & \textbf{Small} &
\textbf{All sizes} & \textbf{Small}\\
\midrule
\multirow{6}{*}{AT} 
& YOLO11m       & 20.1 & 0.093 & 0.012 & 0.040 & 0.005 & 0.104 & 0.022 & 0.050 & 0.001 \\
& YOLO11x       & 56.9 & 0.124 & 0.029 & 0.060 & 0.009 & 0.148 & 0.034 & 0.093 & 0.007 \\
& DETR-Med      & 41.3 & 0.092 & 0.035 & 0.031 & 0.003 & 0.092 & 0.021 & 0.043 & 0.000 \\
& DETR-Large    & 60.0 & 0.066 & 0.026 & 0.035 & 0.008 & 0.104 & 0.025 & 0.057 & 0.002 \\
& LWDETR-Tiny   & 12.1 & 0.158 & 0.068 & 0.057 & 0.000 & 0.136 & 0.039 & 0.062 & 0.003 \\
& LWDETR-Med    & 28.2 & 0.192 & 0.061 & 0.081 & 0.017 & 0.164 & 0.049 & 0.069 & 0.004 \\
\hline
\multirow{6}{*}{DATUM} 
& YOLO11m       & 5.8+20.1 & 0.307 & 0.057 & 0.151 & 0.033 & 0.276 & 0.130 & 0.122 & 0.012 \\
& YOLO11x       & 5.8+56.9 & 0.317 & 0.060 & 0.166 & 0.032 & 0.291 & 0.132 & 0.133 & 0.013 \\
& DETR-Med      & 5.8+41.3 & 0.218 & 0.076 & 0.108 & 0.034 & 0.195 & 0.074 & 0.091 & 0.009\\
& DETR-Large    & 5.8+60.0 & 0.243 & 0.100 & 0.120 & 0.051 & 0.216 & 0.087 & 0.102 & 0.007\\
& LWDETR-Tiny   & 5.8+12.1 & 0.293 & 0.158 & 0.190 & 0.074 & 0.265 & 0.120 & 0.113 & 0.011\\
& LWDETR-Med    & 5.8+28.2 & 0.369 & \underline{0.200} & 0.231 & 0.098 & 0.306 & 0.140 & 0.154 & 0.015\\
\hline
\multirow{6}{*}{TMT} 
& YOLO11m       & 22.9+20.1 & 0.276 & 0.042 & 0.154 & 0.028 & 0.242 & 0.092 & 0.103 & 0.006\\
& YOLO11x       & 22.9+56.9 & 0.307 & 0.038 & 0.168 & 0.026 & 0.256 & 0.092 & 0.112 & 0.008 \\
& DETR-Med      & 22.9+41.3 & 0.204 & 0.062 & 0.101 & 0.029 & 0.162 & 0.045 & 0.076 & 0.002 \\
& DETR-Large    & 22.9+60.0 & 0.224 & 0.087 & 0.129 & 0.040 & 0.178 & 0.052 & 0.093 & 0.003\\
& LWDETR-Tiny   & 22.9+12.1 & 0.297 & 0.134 & 0.188 & 0.074 & 0.237 & 0.097 & 0.101 & 0.005\\
& LWDETR-Med    & 22.9+28.2 & 0.344 & 0.165 & 0.226 & 0.088 & 0.271 & 0.119 & 0.126 & 0.007\\
\hline
\multirow{6}{*}{MAMBA} 
& YOLO11m       & 2.8+20.1 & 0.312 & 0.046 & 0.177 & 0.030 & 0.285 & 0.130 & 0.134 & 0.013 \\
& YOLO11x       & 2.8+56.9 & 0.350 & 0.037 & 0.191 & 0.031 & 0.292 & 0.128 & 0.147 & 0.011\\
& DETR-Med      & 2.8+41.3 & 0.227 & 0.073 & 0.113 & 0.031 & 0.200 & 0.067 & 0.098 & 0.006 \\
& DETR-Large    & 2.8+60.0 & 0.275 & 0.097 & 0.135 & 0.041 & 0.217 & 0.076 & 0.128 & 0.008 \\
& LWDETR-Tiny   & 2.8+12.1 & 0.316 & 0.164 & 0.193 & 0.076 & 0.270 & 0.115 & 0.135 & 0.012 \\
& LWDETR-Med    & 2.8+28.2 & 0.364 & 0.186 & \textit{0.249} & 0.105 & 0.310 & 0.138 & 0.167 & 0.017 \\
\hline
\multicolumn{2}{c|}{DMAT-Tiny (ours)}  & 2.8+12.1 & \textit{0.374} & \textit{0.188} & 0.214 & \textit{0.108} & \textit{0.348} & \textit{0.178} & \textit{0.214} & \textit{0.127} \\
\multicolumn{2}{c|}{DMAT-Small  (ours)} & 2.8+14.6 & \underline{0.407} & \textbf{0.234} & \underline{0.276} & \underline{0.134} & \underline{0.373} & \underline{0.194} & \underline{0.251} & \underline{0.143} \\
\multicolumn{2}{c|}{DMAT-Med (ours)}   & 2.8+28.2 &
\textbf{{0.419}} &
\textbf{{0.234}} &
\textbf{{0.282}} &
\textbf{{0.137}} &
\textbf{{0.385}} &
\textbf{{0.199}} &
\textbf{{0.269}} &
\textbf{{0.152}}\\
\bottomrule
\end{tabular}}
\caption{Object detection performance in terms of mAP[0.50-0.95] for AT mitigation methods and detectors across COCO subsets and object sizes. The top three results are highlighted in \textbf{bold}, \underline{underline}, and \textit{italic}, respectively.  Parameters column shows number of parameters in the object detector and (where applicable) the number of parameters in the AT-mitigator}
\label{tab:method_grouped_mAP_final}
\end{table*}

\subsubsection{Static scenes} 
We use the COCO2017 dataset~\cite{lin2014microsoft}, which includes over 330k images, with 200k labeled for object detection across 80 categories and over 5 million annotated instances. It is divided into training (118k images), validation (5k images), and test (41k images) sets; since test annotations are unavailable, we use the validation set for evaluation.
To explore different use cases, we define three application subsets:

\noindent i) \textbf{All}: All 80 COCO categories;

\noindent ii) \textbf{Top10}: The 10 most frequent categories—Person, Car, Chair, Book, Bottle, Cup, Dining Table, Bowl, Skis, and Handbag;

\noindent iii) \textbf{CarPerson}: Only the Car and Person classes, relevant to surveillance tasks.

For each image, we generate 50-frame turbulent video sequences, with corresponding ground truth videos comprising 50 identical frames. The final datasets include 4,594 training and 503 validation videos for All, 5,000 training and 668 validation videos for Top10, and 5,000 training and 831 validation videos for CarPerson.

\subsubsection{Dynamic scenes}
We use the GOT-10k dataset~\cite{huang2021got10k}, which contains over 10k video segments with bounding box annotations under real-world conditions. For our purposes, we select sequences containing Person or Car objects, extract 50-frame clips and apply synthetic distortions to simulate the spatial and temporal characteristics of AT. As the GOT-10k dataset labels only one object per frame, we generate pseudo ground truth using the YOLOv11x detector. In total, 1,761 videos are used for training and 441 for testing. 

\subsection{Experiment settings}

We compare our proposed DMAT with three state-of-the-art AT mitigation methods: TMT~\cite{Zhang:Image:2024}, DATUM~\cite{zhang2024spatio}, and MAMBA~\cite{gong2024nnmamba}, as well as three object detectors: YOLOv11~\cite{khanam2024yolov11}, DETR~\cite{carion2020end}, and LW-DETR~\cite{chen2024lw}, each evaluated at multiple model sizes. Note that we do not apply joint optimization to the baseline methods, as doing so would incorporate our contributions into their architectures, resulting in an unfair comparison.

All models are fine-tuned on the new synthetic datasets described above. All models use the Adam optimizer with a learning rate of 0.0001 for 100 epochs, without early stopping. A sliding window approach was used during both training and inference, where each target frame was predicted from ten temporally adjacent frames (five preceding, four following), following the setup in~\cite{anantrasirichai2023atmospheric}, which balances performance and temporal context. The implementation was carried out in Python, utilizing the PyTorch framework with CUDA acceleration to ensure computational efficiency. All experiments were conducted on NVIDIA 4090 and A100 GPUs.

\section{Results and discussion}
\label{sec:results}

We evaluate system performance in both object detection and atmospheric turbulence (AT) mitigation using our proposed synthetic datasets and real AT videos. For synthetic datasets, where ground truth is available, we report quantitative metrics, while real AT videos allow only for subjective visual assessment. Object detection performance is measured using Average Precision (AP) across multiple Intersection-over-Union (IoU) thresholds (0.50 to 0.95 in 0.05 increments, denoted AP[0.50–0.95]). We report Mean Average Precision (mAP), which averages AP across all object categories. For AT mitigation, we use Peak Signal-to-Noise Ratio (PSNR), Structural Similarity Index Measure (SSIM)~\cite{wang2004image}, and the perceptual similarity metric LPIPS~\cite{zhang2021perceptual}.

\begin{figure}
    \centering
    \includegraphics[width=1\linewidth]{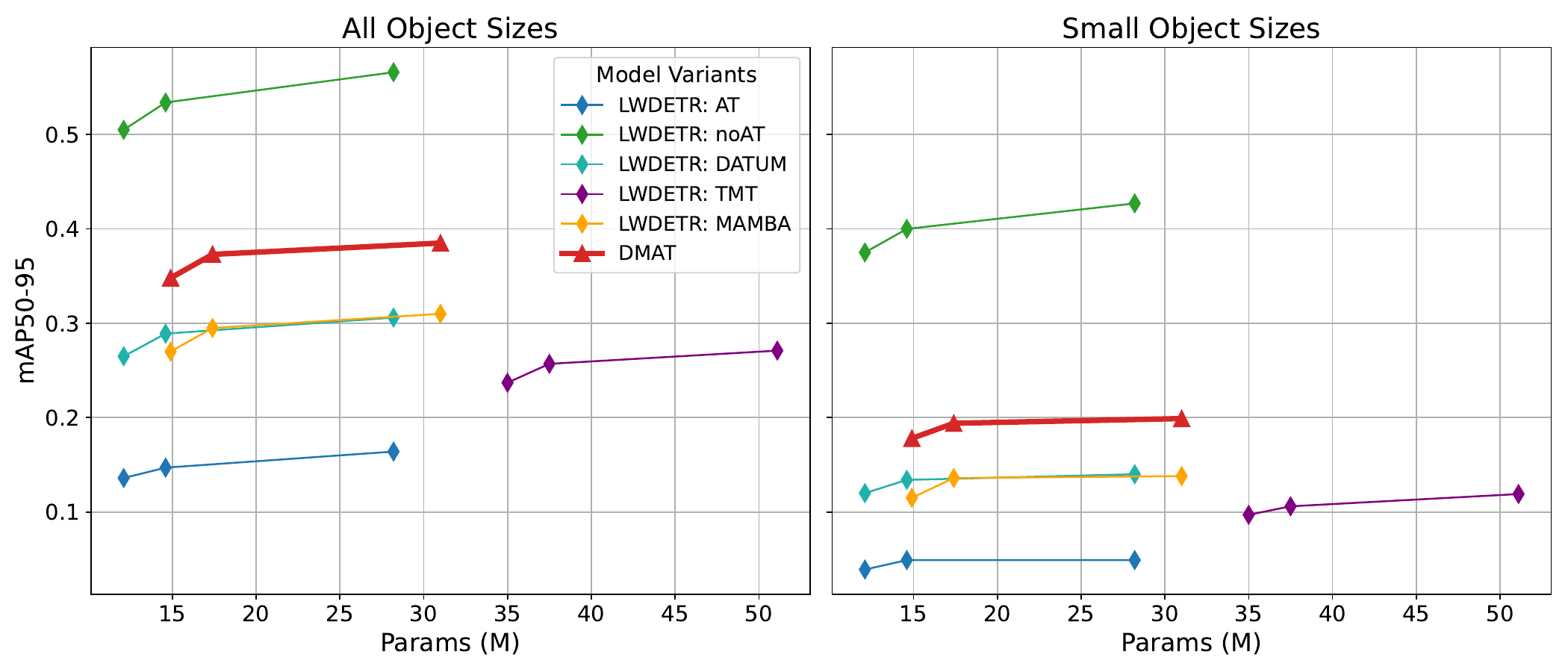} \\
    (a) Static scenes with car and Person \\
    \includegraphics[width=1\linewidth]{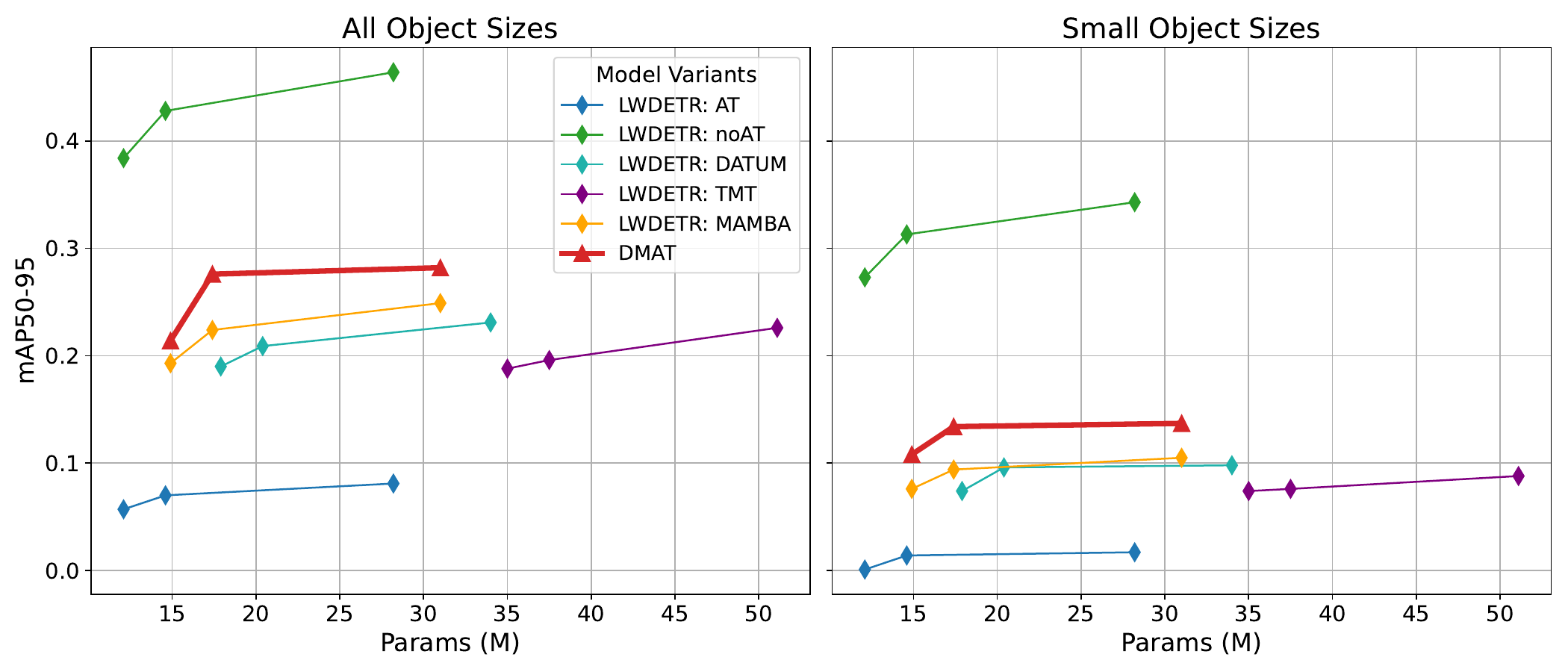} \\
    (b) Static scenes with top 10 objects \\
    \includegraphics[width=1\linewidth]{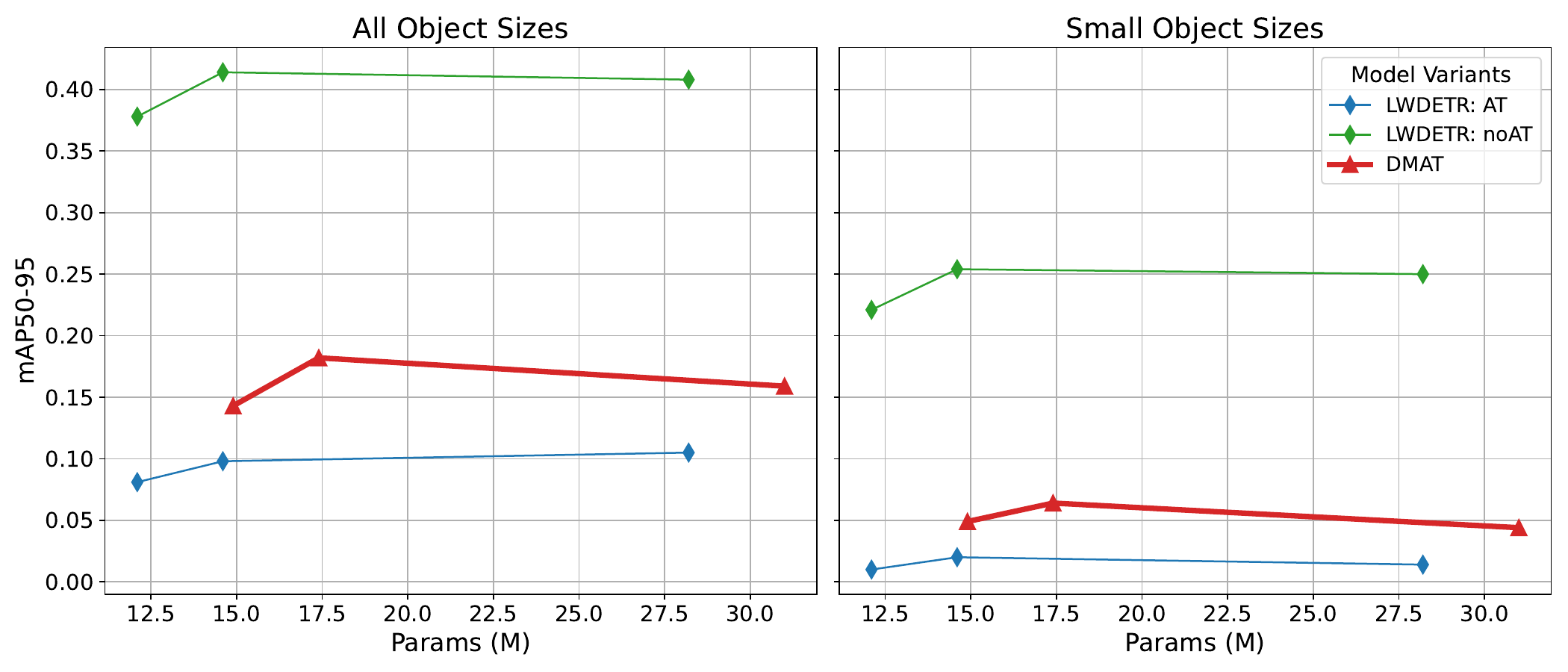} \\
    (c) Dynamic scenes with car and Person \\
    \caption{Object detection performance (mAP[0.50-0.95]) for various object class groupings across all (left) and small (right) object sizes. 
 Number of parameters is the size of the object detector only (for AT and noAT) and the sum of parameters of the AT mitigation model and the object detector.}
    \label{fig:top_combined}
\end{figure}

\subsection{Synthetic datasets}

\subsubsection{Performance of object detection.}
The results presented in Table~\ref{tab:method_grouped_mAP_final} demonstrate that our proposed DMAT framework outperforms existing methods in object detection under AT conditions on COCO datasets. Among two-stage approaches employing the same object detector, MAMBA achieves higher performance than TMT and DATUM. However, compared to our proposed DMAT, its performance can be up to 25\% lower when using models of similar size. This highlights the effectiveness of our joint framework, where the AT mitigator generates restored frames with semantically enriched features that are more conducive to accurate object detection.

Fig.~\ref{fig:top_combined} shows the object detection performance of our DMAT model compared with various baseline models, applied either independently or sequentially, on the COCO and GOT-10k datasets. The graphs include results for non-mitigated ground truth inputs (noAT) and turbulent inputs (AT), serving as upper- and lower-bound benchmarks, respectively. We present results for 2 and 10 object classes; performance across all object sizes and classes follows a similar trend, as shown in Fig.~\ref{fig:adresults}. The results highlight two key trends: (i) detection performance generally improves with increasing model size; and (ii) the Mamba-based AT mitigation model consistently outperforms both DATUM and TMT across all cases.

\begin{figure}
    \centering
    \includegraphics[width=1.05\linewidth, trim=0 0 0 9.6cm, clip]{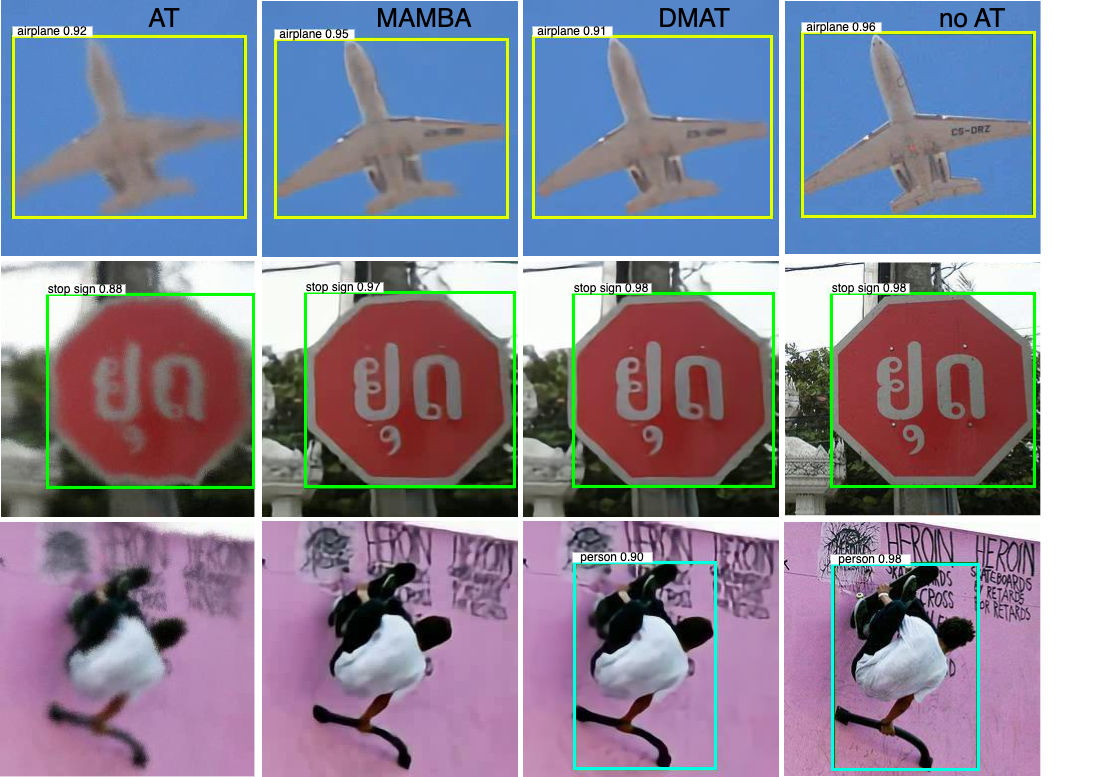}
    \caption{Comparison of AT-mitigation and Object Detection Methods.  All methods apart from DMAT use the medium sized LW-DETR method whereas the DMAT method uses the Medium sized architecture.}
    \label{fig:montage2}
\end{figure}

The right hand side of Fig.~\ref{fig:top_combined}  display the same set of methods evaluated specifically on small objects (defined as those smaller than $32\times32$ pixels). These graphs further illustrate that: (i) for all methods, mAP[0.50-0.95] degrades significantly when limited to small objects, which is expected due to the increased blurring effect of turbulence on smaller targets, even when AT-mitigation is applied; and (ii) the DMAT models maintain a performance advantage over individual methods, consistent with the trends observed across all object sizes. Fig.~\ref{fig:montage2} shows some example detections.  All the shown methods apart from DMAT use the medium sized LW-DETR method (28.2 Million parameters) whereas the DMAT method uses the Medium sized architecture.  This shows that object detection is obviously difficult for the distorted input (AT), improves with mitigation, and confirms that DMAT achieves superior performance.

\begin{table*}[h!]
\centering
\resizebox{0.95\textwidth}{!}{
\begin{tabular}{l|ccc|ccc|ccc|ccc}
\toprule
\multirow{3}{*}{\textbf{AT Mitigation}} & \multicolumn{9}{c|}{\textbf{Static scenes}} & \multicolumn{3}{c}{\textbf{Dynamic scenes}} \\ \cline{2-13}
& \multicolumn{3}{c|}{\textbf{All}} 
& \multicolumn{3}{c|}{\textbf{Top10}} & \multicolumn{3}{c|}{\textbf{CarPerson}} & \multicolumn{3}{c}{\textbf{CarPerson}}  \\ \cline{2-13}
& PSNR $\uparrow$ & SSIM $\uparrow$ & LPIPS  $\downarrow$ 
& PSNR $\uparrow$ & SSIM $\uparrow$ & LPIPS$\downarrow$ & PSNR $\uparrow$ & SSIM $\uparrow$ & LPIPS$\downarrow$  & PSNR $\uparrow$ & SSIM $\uparrow$ & LPIPS$\downarrow$\\
\midrule
AT & {20.946} & {0.512} & {0.546} 
   & {20.668} & {0.513} & {0.550}
   & {20.210} & {0.488} & {0.564}
   & {22.252} & {0.591} & {0.466} \\
DATUM 
   & 23.092 & 0.644 & \underline{0.356} 
   & 22.910 & 0.649 & \underline{0.343} 
   & 22.390 & 0.632 & \underline{0.360}
   & 22.564 & 0.693 & \textit{0.373}\\
TMT 
   & \textit{23.332} & \textit{0.654} & \textbf{{0.348}} 
   & \textit{23.167} & \textit{0.661} & \textbf{{0.331}}
   & \textit{22.592} & \textit{0.641} & \textbf{0.350} 
   & \textit{22.747} & \textit{0.712} & \textbf{0.367}\\
MAMBA 
   & \underline{23.763} & \underline{0.667} & 0.378 
   & \underline{23.693} & \underline{0.677} & 0.360 
   & \underline{23.084} & \underline{0.657} & 0.380 
   & \underline{23.297} & \underline{0.783} & 0.379 \\ \hline
DMAT (ours)
   & \textbf{{23.841}} & \textbf{{0.671}} & \textit{{0.373}} & \textbf{{23.861}} & \textbf{{0.683}} & \textit{0.357} 
   & \textbf{23.220} & \textbf{0.663} & \textit{0.376}  & \textbf{23.312} & \textbf{0.831} & \underline{0.371} \\
\bottomrule
\end{tabular}}
\caption{Comparison of AT Mitigation methods for synthetic static scenes across two datasets: All and Top10 categories. Best values are highlighted in \textbf{bold}, \underline{underline}, and \textit{italic}.}
\label{tab:combined_comparisonpsnretc}
\end{table*}

\subsubsection{Performance of AT mitigation.}

Table~\ref{tab:combined_comparisonpsnretc} shows objective comparison results of each AT mitigation method (compared to the ground truth no-AT sequences).  This table shows that the DMAT method gives the best performance in terms of distortion measured by PSNR and SSIM.   This superior performance may be attributed to its additional training stages that explicitly incorporate a distortion loss. Alternatively, or in conjunction, the enhanced results could stem from the joint optimization of the object detection loss, which likely contributes to improved structural fidelity in the reconstructed objects.

\begin{figure}
    \centering
    \includegraphics[width=1\linewidth]{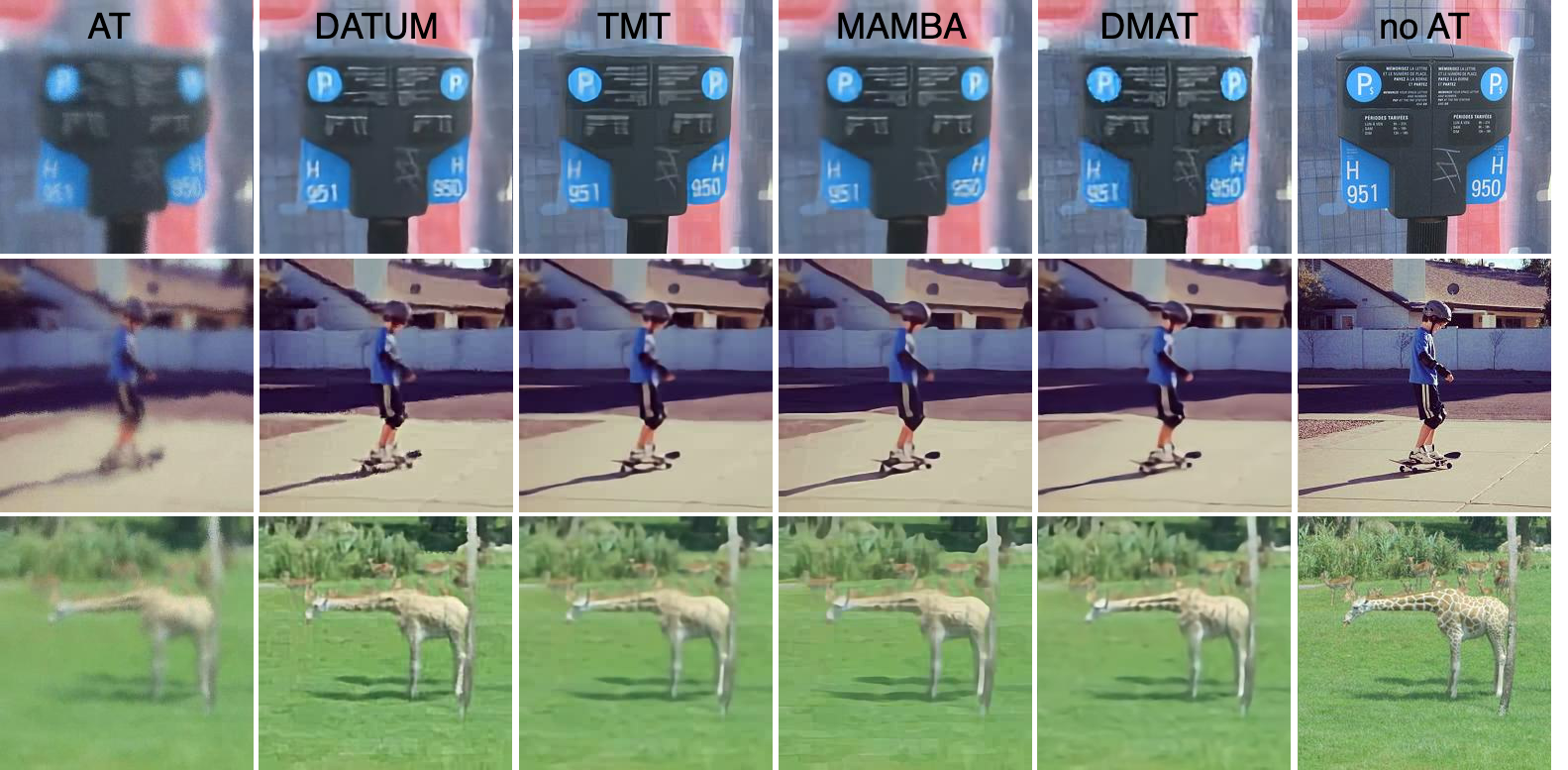}
    \caption{Comparison of AT-mitigation methods}
    \label{fig:montage1}
\end{figure}

Fig.~\ref{fig:montage1} presents a subjective comparison of the results from DMAT compared with TMT, DATUM and MAMBA methods. Subjectively, DATUM preserves textures the best, but turbulence effects remain at the object edges. TMT produces sharper edges but at the expense of texture loss. The MAMBA method strikes a balance between the two, offering a compromise on texture preservation and edge definition. When incorporating the object detector, our DMAT method enhances sharpness around foreground objects but sacrifices texture details in the background compared to the MAMBA method.

\subsection{Real AT sequences}
\begin{figure*}[h]
    \centering
    \includegraphics[width=\linewidth]{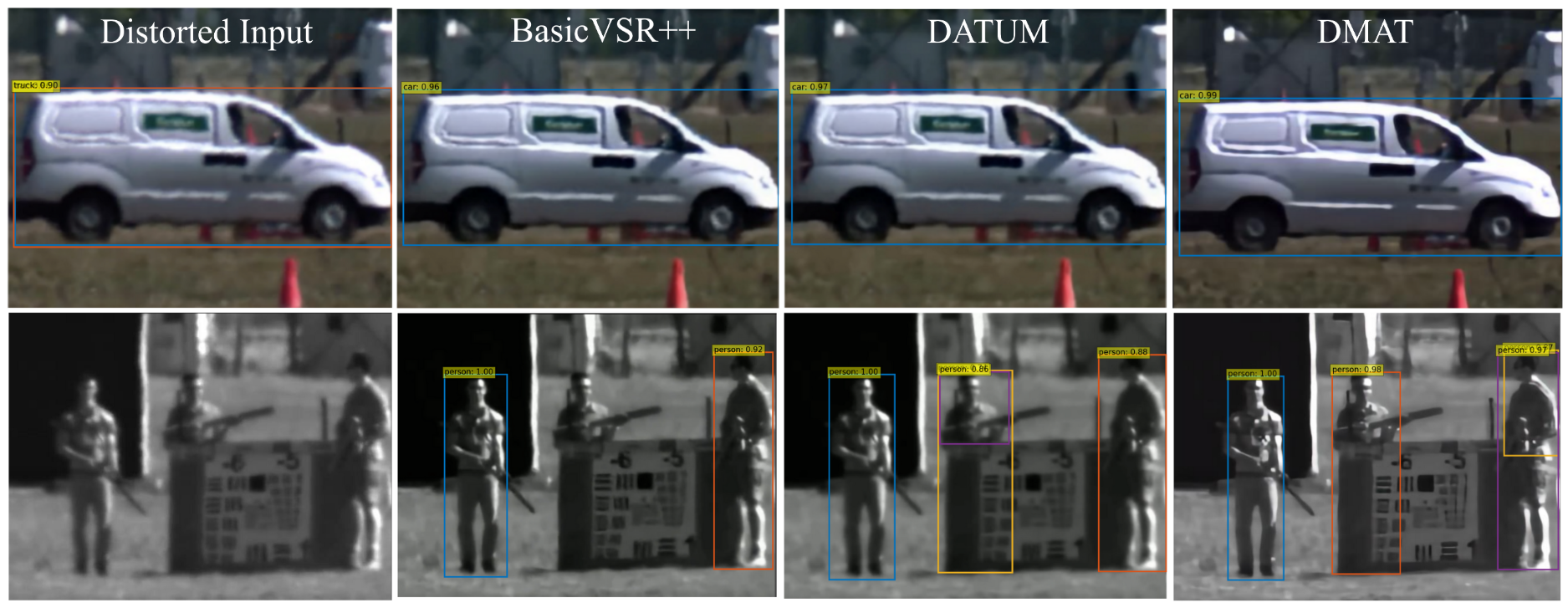}
    \caption{Qualitative detection results on real-world atmospheric turbulence videos from the \textbf{CLEAR} dataset. From left to right: distorted input, BasicVSR++, DATUM, and our proposed DMAT. All baseline methods adopt the medium-sized LW-DETR method, while DMAT uses its own medium-sized configuration. Zoom in for better visualization.}.
    \label{fig:realAT}
\end{figure*}
To further verify the generalisation ability of our proposed DMAT framework in real atmospheric turbulence scenarios, we tested it on real turbulent videos from the CLEAR dataset \cite{Anantrasirichai:Atmospheric:2013}. These videos have no ground truth labels, so we can only make visual comparisons. Fig.~\ref{fig:realAT} shows a qualitative comparison between DMAT and two typical restoration methods: BasicVSR++~\cite{chan2022basicvsrpp} and DATUM~\cite{zhang2024spatio}. To keep the comparison fair, all baseline methods used the medium-size LW-DETR detector, while DMAT used its own medium-size version.

In the van scene (top row), where the van moves from left to right, the raw input has strong geometric distortion and edge blur, leading to very low detection confidence. BasicVSR++ reduces some blur but still gives distorted object shapes and unstable detection. DATUM gives clearer object outlines than BasicVSR++, but there are still turbulence artefacts around the edges, which affect detection. In contrast, DMAT removes most distortions, restores sharper and more complete vehicle structures (like wheels and windows), and gives much higher detection confidence.

In the human scene (bottom row), where three men are moving objects in their hands, the raw input shows flickering and motion ghosting, so the detector often misses the targets. BasicVSR++ partly restores the human shapes, but detection is still unstable. DATUM detects several targets but with low confidence scores. In contrast, DMAT restores clear and complete human shapes, allowing the detector to find all targets with high confidence. This shows that DMAT generalises well from synthetic data to real turbulent conditions.

Overall, DMAT can keep object structures clear and consistent under real turbulence, while traditional restoration-only methods cannot achieve this.


\subsection{Ablation study}
\textbf{Experiment 1:} As seen in Table~\ref{tab:method_grouped_mAP_final} and Fig.~\ref{fig:top_combined}, when the AT mitigator is removed, the performance of object detection drops by approximately 50\% on average, with an even larger drop for smaller objects. 

\noindent \textbf{Experiment 2:} When the object detection module is removed, the performance of AT mitigation drops by up to approximately 1.5\% in term of PSNR. 

\noindent \textbf{Experiment 3:} When the feature sharing between the two modules is removed (red lines in~\cref{fig:system_architecture}), the performance degradation is reduced by up to 1\% in terms of PSNR and 2\% in terms of mAP, indicating that better integration of features contributes to both improved reconstruction quality and detection accuracy.

\section{Conclusion}
We propose the first end-to-end framework that jointly tackles atmospheric turbulence (AT) mitigation and object detection. By integrating a 3D Mamba-based restoration module with a transformer-based detector, our architecture enables mutual enhancement between low-level and high-level features. This improves detection accuracy under severe distortions while preserving object-level details. Experiments on turbulent COCO datasets show our method outperforms separate AT-mitigation and detection models, achieving up to a 15\% mAP[0.50–0.95] gain. Bidirectional feature flow improves robustness, especially for small and surveillance-relevant objects. While performance is strong, the joint model requires more memory during training due to its combined architecture. 
\small
\pagebreak
\bibliographystyle{plain}
\bibliography{main,literature_review, sim_lit,sn-bibliography}
\end{document}